\documentclass[letterpaper, 10 pt, conference]{ieeeconf} 
\IEEEoverridecommandlockouts 
\usepackage{cite}
\usepackage{amsmath,amssymb,amsfonts}
\usepackage[super]{nth}
\usepackage{algorithmic}
\usepackage{xcolor}
\usepackage{graphicx}
\usepackage{textcomp}
\usepackage{xcolor}
\usepackage{graphics} 
\usepackage{epsfig}
\usepackage{xcolor}
\usepackage{nicefrac}
\usepackage{siunitx}
\usepackage{numprint}
\usepackage{url}
\usepackage{ragged2e}
\usepackage{bbding}
\def\BibTeX{{\rm B\kern-.05em{\sc i\kern-.025em b}\kern-.08em
    T\kern-.1667em\lower.7ex\hbox{E}\kern-.125emX}}
\newcommand{\railgoerl}{RailGoerl24}
\makeatletter 
\newcommand{\linebreakand}{%
  \end{@IEEEauthorhalign}
  \hfill\mbox{}\par
  \mbox{}\hfill\begin{@IEEEauthorhalign}
}
\makeatother 
\usepackage{hyperref}
\hypersetup{
    colorlinks=true,
    linkcolor=blue,
    filecolor=blue,      
    urlcolor=blue,
    pdftitle={\railgoerl},
    pdfpagemode=FullScreen,
    }

\title{\LARGE \bf \railgoerl: Görlitz Rail Test Center CV Dataset 2024}

\author{Rustam Tagiew$^{1}$, Ilkay Wunderlich$^{2}$, Mark Sastuba$^{1}$, Kilian Göller$^{3}$ and Steffen Seitz$^{3}$ 
\thanks{\tiny $^{1}$Rustam Tagiew and Mark Sastuba are with the German Centre for Rail Traffic Research at the Federal Railway Authority (DZSF), Dresden, Germany. This is not an official statement, guideline or directive of the German Federal Railway Authority.
{\tt forschung@dzsf.bund.de}}%
\thanks{\tiny $^{2}$Ilkay Wunderlich is with EYYES GmbH, Gedersdorf, Austria
        {\tt ilkay.wunderlich@eyyes.com}}%
\thanks{\tiny $^{3}$Steffen Seitz and Kilian Göller  are with  the Chair of Fundamentals of Electrical Engineering of Dresden University of Technology, Dresden, Germany. Steffen Seitz is also with the Conrad Zuse School of Embedded Composite AI (SECAI).
        {\tt steffen.seitz@tu-dresden.de}}
\thanks{\tiny \textcopyright 2023 IEEE. Personal use of this material is permitted. Permission from IEEE must be obtained for all other uses, in any current or future media, including reprinting/republishing this material for advertising or promotional purposes, creating new collective works, for resale or redistribution to servers or lists, or reuse of any copyrighted component of this work in other works. DOI: \href{https://ieeexplore.ieee.org/document/11135724}{: 10.1109/ERAS63351.2025.11135724}}        
}

\begin{document}
\maketitle
\thispagestyle{empty}
\pagestyle{empty}

\begin{abstract}
Driverless train operation for open tracks on urban guided transport and mainline railways requires, among other things automatic detection of actual and potential obstacles, especially humans, in the danger zone of the train's path. Machine learning algorithms have proven to be powerful state-of-the-art tools for this task. However, these algorithms require large amounts of high-quality annotated data containing human beings in railway-specific environments as training data. Unfortunately, the amount of publicly available datasets is not yet sufficient and is significantly inferior to the datasets in the road domain. Therefore, this paper presents \railgoerl, an on-board visual light Full HD camera dataset of \numprint{12205} frames recorded in a railway test center of TÜV SÜD Rail, in Görlitz, Germany. Its main purpose is to support the development of driverless train operation for guided transport. \railgoerl~also includes a terrestrial LiDAR scan covering parts of the area used to acquire the RGB data. In addition to the raw data, the dataset contains \numprint{33556} boxwise annotations in total for the object class `person'. The faces of recorded actors are not blurred or altered in any other way. \railgoerl, available at \href{https://data.fid-move.de/dataset/railgoerl24}{data.fid-move.de/dataset/railgoerl24}, can also be used for tasks beyond collision prediction.
\end{abstract}

\section{INTRODUCTION}
\label{sec:intro}
\begin{figure*}[ht!]
  \centering
\includegraphics[width=1\linewidth]{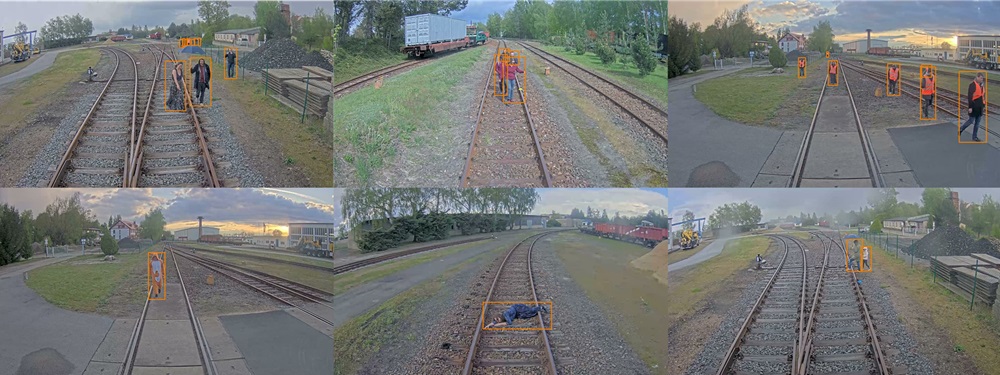}
 \caption{Examples of recorded and annotated scenarios from \railgoerl. The top left and center corners show several persons without warning jackets. The top right corner shows several persons wearing warning jackets. In the bottom left corner is a person in a kangaroo costume. The bottom center presents a person lying on the tracks and the bottom right corner shows a person with a child. Annotating rectangles are orange.}
  \label{overview}
\end{figure*}
In automatic train operation (ATO), technical systems replace the functions of the train driver and other operating staff. ATO includes different grades of automation (GoA) \cite{spec99002}. The higher the GoA, the more functions are replaced, up to GoA4 with no staff on-board. At least in the European context, regulations and technical standards differ significantly for mainline railways and urban guided transport like subways, light rail, and trams. In the case of GoA2, this leads to the parallel development of similar systems such as the European Train Control System (ETCS) for mainline railways and Communication-Based Train Control (CBTC) for urban systems. However, for the person detection function from GoA3 onwards, it is expected that related system components will be developed by applying machine learning (ML) to shared datasets. 
 
Monitoring the danger zone of the train's path is one of the main tasks of the train driver, who predicts collisions and acts accordingly. This task is the most challenging as it impedes an upgrade to GoA3 and GoA4 (GoA3+) where no driver is required \cite{osdar23}. In this regard, GoA3+ of mainline trains differs from fully automated metros, such as the Nuremberg U-Bahn. They are also known as ``horizontal lifts'', closed systems with traffic running in isolated and mostly enclosed environments, such as tunnels. As a result, no person detection is needed and, consequently, no computer vision (CV) systems are required onboard. In contrast, open mainline and open urban rail systems are subject to environmental interferences, e.g., at level crossings or on unfenced tracks. 

The CV dataset \railgoerl\cite{railgoerl24}, presented in this paper, is proposed to facilitate the ML-based development of onboard person detection systems, as well as stationary automated security surveillance and occupational safety systems. Additionally, the few currently available related real-world datasets including objects in the railway environment are listed in Sec.\ref{rvd}.

\section{Existing Datasets}
\label{rvd}
To the best of the authors' knowledge, datasets listed in Tab.\ref{exdrv} are the CV datasets recorded by real-world frontal on-board sensors of a mainline or urban railway train including object annotations that have explicitly been published for free use by the research community. Purely synthetic or datasets of artificially injected objects like RailFOD23\cite{chen2024railfod23} and SRLC\cite{srlcdataset} are not included in Tab.\ref{exdrv}. The listed datasets contain annotated single RGB-camera frames from video sequences, LiDAR recordings and rarely recordings from other sensors. 

The datasets RailSem19\cite{railsem19}, RAWPED\cite{rawped} and rail-hp8ij\cite{railhp8ij} contain single RGB camera images of mainline or urban railway scenes with polygonal annotations of persons. RailSem19\cite{railsem19} additionally provides dense pixel-wise semantic segmentation of persons. RailEye3D\cite{Raileye3d} is a stereo RGB HD dataset containing polygonal annotations of persons on platforms mainly for the task of safe door operation. ESRORAD\cite{khemmar2022road} is a multi-sensor dataset recorded by a car driving on paved tram rails. RailVID\cite{yuan2022railvid} is an open polygonal annotated infrared dataset for urban railways. The first open annotated multi-sensor dataset for mainline railways is OSDaR23\cite{osdar23}. RailEye3D\cite{Raileye3d}, ESRORAD\cite{khemmar2022road}, RailVD\cite{yuan2022railvid} and OSDaR23\cite{osdar23} contain cuboid and polygonal annotations of persons.

In summary, there are only $7$ relatively small annotated open CV datasets for the class `person' from the years 2019--2023. The growth of their number is low and not significantly increasing over the last years.
\section{Motivation for \railgoerl}
\label{propdataset}
The danger zone of a train can be entered with or without suicidal intent. Suicidal intent is excluded from the safety assessment by EN 62267 \cite{en62267} and CSM-RA\cite{csmra} - only accidents are considered. It is therefore not mentioned in the safety requirements. However, it should be noted that suicidal intent can only be determined after the incident, if at all. Therefore, this difference is irrelevant for person detection developers. \railgoerl~was designed to contain also scenes with certain suicidal appearances, some of which are shown in Fig.\ref{overview}. 
\section{Choice of Scenarios}
\label{scenarios}
One of the main goals in the choice of scenarios for \railgoerl~was achieving representative diversity of recorded human beings for a railway operational design domain in Germany. The diversity is defined in multiple dimensions. The human amateur actors were recorded with and without the obligatory warning jackets. Male and female actors were involved, their age varied between children and elderly. Variations of medical conditions included walking disability and pregnancy. Clothing styles, body positions and crowd sizes were varied as well. The desired diversity makes our dataset ideal for evaluating trained ML models since train-related edge cases (e.g. a person lying on the tracks) are not represented in previous open datasets - yet they could still occur in practice. 

The most important concern was the safety of the human actors. To prevent accidents, the train was always moved away from human actors, never towards them. All recordings took place on April 24th, 2024.

Additionally, human actors signed an agreement that their faces would not be altered due to data protection reasons in the final dataset publication. This prevents inducing possible biases in the data, which could be exploited by "Clever Hans" predictors. Such a "Clever Hans" predictor otherwise learns to detect altered areas in the frames instead of human features, as all pictures containing humans are altered as well. Explainable artificial intelligence (XAI) methods would detect such a predictor. They keep humans in the loop by enabling verification of ML algorithms by human experts via providing saliency maps \cite{XAI}.
\begin{table*}
  \centering
  \caption{Existing real-world sensor data in open datasets for mainline and urban GoA3+}
  \begin{tabular}{|l||r||r||r||r||r||r||r||r|}
    \hline
    \textbf{Dataset}& \textbf{Pub.} & \textbf{Size (frs.--frames,} & \textbf{Sensors} & \textbf{Data format} & \textbf{Annotation} &\textbf{Main-} & \textbf{Urban} & \textbf{Class}\\

           & \textbf{year}& \textbf{pts.-- 3D points)}& &  & & \textbf{line}& & \textbf{`person'}\\
    \hline
     RailSem19\cite{railsem19}    & 2019 & $\numprint{8500}$ frs.   & RGB & variable                     & polygon,          & \checkmark & \checkmark & \checkmark \\
                                  &      &                          &     &                              & 2D semantic,      &            &            & \\
                                  &      &                          &     &                              & splines           &            &            & \\
     \hline
     FRSign\cite{FRSign}          & 2020 & $\numprint{105352}$ frs. & RGB & $\numprint{2048}\times\numprint{1536}$,           &  polygon          &  \checkmark &     & \\
                                  &      &                          &     & $\numprint{1920}\times \numprint{1200}$            &                   &             &     & \\
     \hline
     RAWPED\cite{rawped}          & 2020 & $\numprint{26000}$ frs.  & RGB & variable                     & polygon           &  \checkmark & \checkmark & \checkmark\\ 
     \hline
     Catenary Arch\cite{catenaryarch}         & 2021 & $\numprint{55}\times10^{6}$ pts.   & LiDAR &             & 3D semantic       &  \checkmark &     & \\
     \hline
     RailEye3D\cite{Raileye3d}    & 2021      &  $\numprint{11867}$ frs.                       & $2\times$RGB            &  $\numprint{1088}\times\numprint{1920}$  & polygon                   & \checkmark &      & \checkmark \\
     \hline
     Rail-DB\cite{raildb}         & 2022 & $\numprint{7432}$ frs.   & RGB & $800\times 288$              &   polyline        &  \checkmark &     & \\
     \hline
     *RailSet\cite{railset}        & 2022 & $\numprint{6600}$ frs.   & RGB & variable                     &  polygon,       & \checkmark  &     & \\
         &   &   &   &  &  polyline                  &         & & \\
     \hline
     *ESRORAD\cite{khemmar2022road}& 2022 & **$\numprint{100000}$ frs.  & $2\times$RGB, LiDAR &                      $\numprint{1920}\times\numprint{1080}$  &    cuboid     &  &  \checkmark    & \checkmark\\
     \hline
     RailVID\cite{yuan2022railvid} & 2022 & $\numprint{1071}$ frs.  & Infrared   & $640\times 512$                       & 2D semantic       &  & \checkmark     & \checkmark\\
     \hline
     rail-hp8ij\cite{railhp8ij}                     &  2022  &            $578$ frs.               & RGB            &    variable          &    polygon  & \checkmark &      & \checkmark\\
     \hline
     GERALD\cite{gerald}          & 2023 & $\numprint{5000}$ frs.  & RGB        & $\numprint{1920}\times\numprint{1080}$,      & polygon     & \checkmark &      & \\
                                  &      &                         &            & $\numprint{1280}\times\numprint{720}$     &                   &            &      & \\
     \hline
     OSDaR23\cite{osdar23}        & 2023 & $\numprint{1534}$ m-frs.                     & $6\times$RGB,  &   $\numprint{4112}\times\numprint{2504}$,                 & polygon,          & \checkmark &      & \checkmark\\
                                  &      &                         &  & $\numprint{2464}\times\numprint{1600}$,                   & cuboid,       &            &      &  \\
                                  &      &                         & $3\times$Infrared, & $\numprint{640}\times\numprint{480}$,                   &    polyline  &           &      &  \\
                                  &      &                         & Radar, & $\numprint{2856}\times\numprint{1428}$                   &     &            &      &  \\
                                  &      &                        & LiDAR &     &    &            &      &  \\
     \hline
     WHU-Railway3D\cite{WHU-Railway3D} & 2023 & $\numprint{4600}\times10^{6}$ pts. & LiDAR &              & 3D semantic       & \checkmark           & \checkmark      & \\
     \hline
     Rail3D\cite{rail3d}          & 2024 & $\numprint{288}\times10^{6}$ pts. & LiDAR &                    & 3D semantic       &  \checkmark          &      & \\

     \hline
     RailPC\cite{railpc}          & 2024 & $\numprint{3000}\times10^{6}$ pts. & LiDAR &                   & 3D semantic       &            &  \checkmark    & \\
     \hline
     RailCloud-HdF\cite{RailCloud-HdF} & 2024 & $\numprint{8060}\times10^{6}$ pts.  & LiDAR &             & 3D semantic       &  \checkmark          &       & \\
     \hline
     \multicolumn{9}{|l|}{*These datasets additionally contain artificially generated data.} \\
     \multicolumn{9}{|l|}{**Not all frames depict urban street running rails; Recorded from a road car making use of the pavement on the rails, not from a railway vehicle.} \\
     \hline
  \end{tabular}
  \label{exdrv}
\end{table*}
\section{Recording Area and Sensor System}
\label{fmss}
\begin{figure}[ht]
  \centering
\includegraphics[width=1\linewidth]{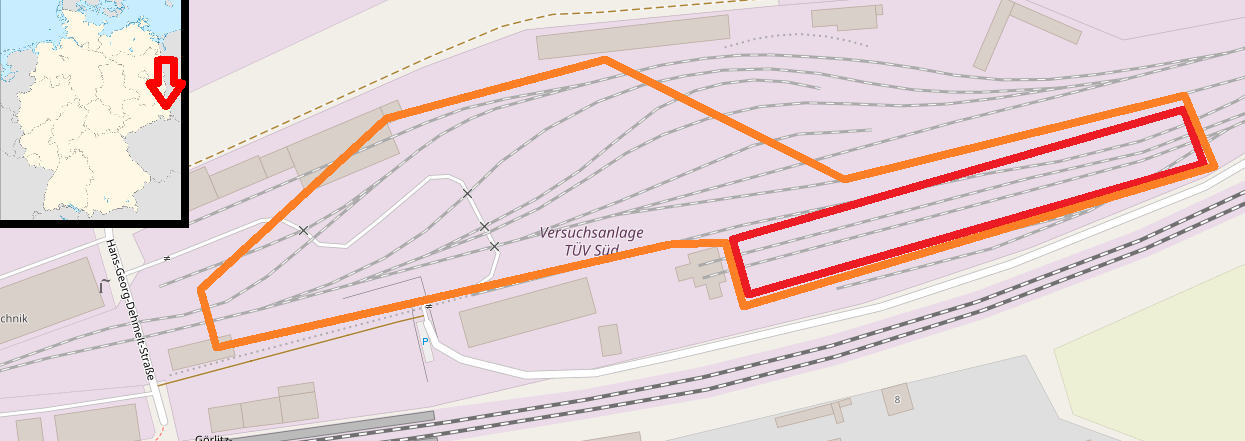}
\caption{Red arrow in the top left map shows the location of TÜV SÜD rail test center in Görlitz, Germany [openstreetmap.org]. The area is marked orange for RGB and red rectangle for the LiDAR recordings.}
 \label{arrea}
\end{figure}

Acquiring access to railway infrastructure to record CV data closest possible to realistic hazardous scenarios is much more complicated than for road-based cases. This has been achieved in the case of \railgoerl~for mainline railways. Fig.\ref{arrea} shows the map of the recorded area near Görlitz, Germany, where trains are operated at GoA0.  

\begin{figure}[ht]
  \centering
\includegraphics[width=1\linewidth]{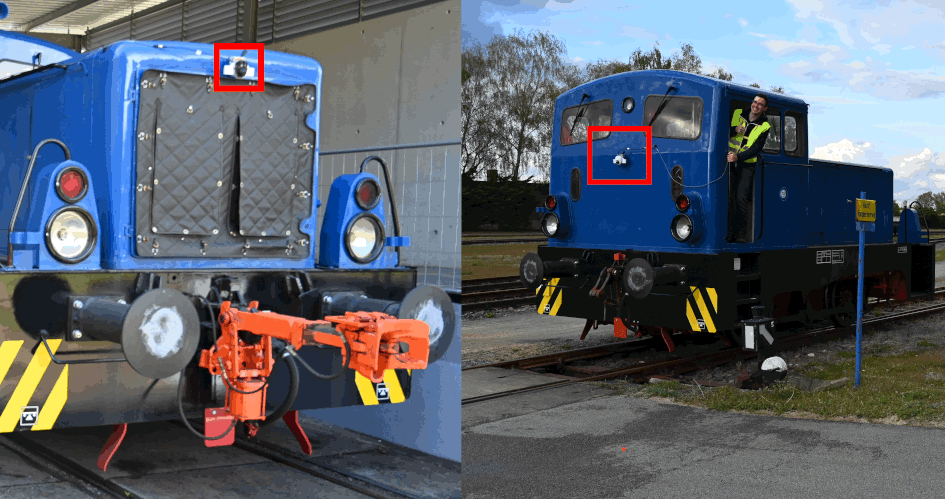 }
\caption{Full HD camera on a V 22 locomotive (SN:\href{https://www.deutsche-kleinloks.de/index.php?nav=1401007&lang=1&id=28947&action=portrait}{0262.6.620}) at $\qty{2.45}{m}$ from the track surface, mounted on the front or rear, marked red.}
 \label{plattform}
\end{figure}

\begin{figure}[ht]
  \centering
\includegraphics[width=0.9\linewidth]{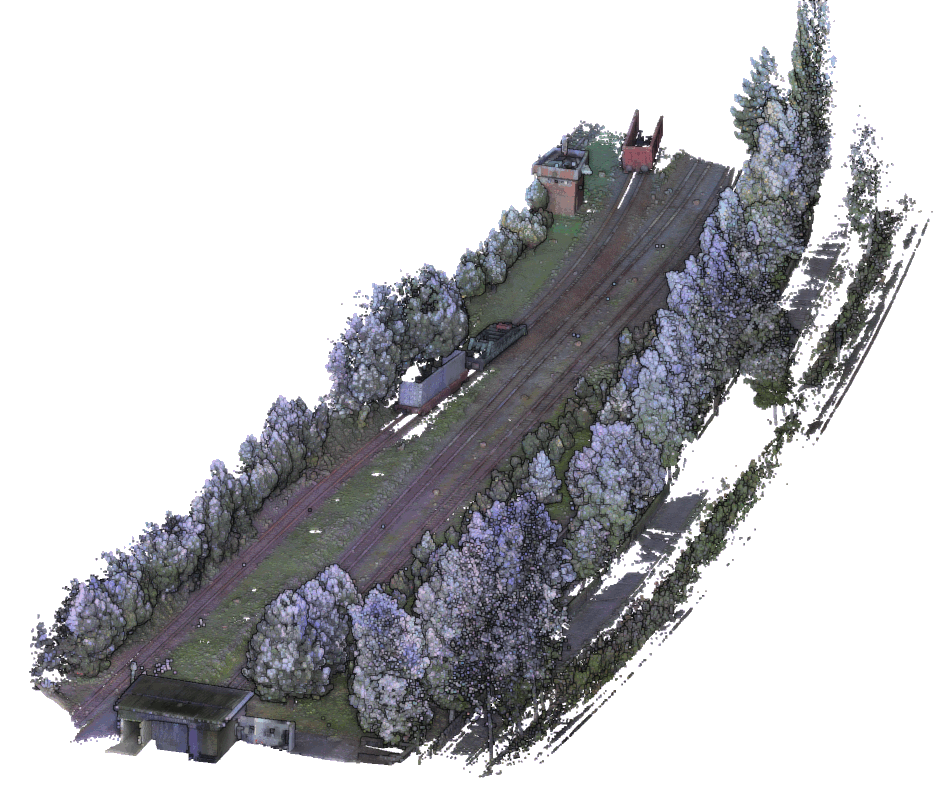}
\caption{LiDAR point cloud of the recorded area, the red rectangle in Fig.\ref{arrea}.}
 \label{pointcloud}
\end{figure}

In Annex I of CSM-RA\cite{csmra} points 2.1.4(b), 2.4.2(b) and 2.4.3(b) allow a simplified approach for safety approval of CV systems for driverless trains, if they have \textit{``similar functions and interfaces''} to the replaced human driver. According to DIN SPEC 91516 \cite{spec91516}, an RGB camera as an interface is more similar to a human eye than a LiDAR sensor. Fig.\ref{plattform} shows the on-board sensor setup with a single RGB $\numprint{1920}\times\numprint{1080}$ camera for recording \railgoerl. It is the model Axis P3925-R with a $\qty{3.6}{mm}$ lens covering a $\qty{85.7}{\text{\textdegree}}$ horizontal and a $\qty{46.0}{\text{\textdegree}}$ vertical field of view. The orientation of the camera is tilted slightly downwards. The bottom edge of the image is the first visible point at a distance of approx. $\qty{3}{m}$ in front of the vehicle.\\
Additionally to RGB, LiDAR data for a share of the surrounding area was acquired. The accompanying 3D data were captured using a Leica RTC360 terrestrial laser scanner and had multiple motivations. It can be used to mimic train driver's route knowledge, create a digital map, employ background subtraction algorithms or simply make a better picture of geometric conditions in the area of recordings.  The recorded colored point cloud consists of $\numprint{383922305}$ points as Fig.\ref{pointcloud} depicts and was merged from $40$ individual scan positions.

In addition to the raw sensor data, \railgoerl~is a semi-automatically annotated by rectangular bounding boxes enclosing all visible persons (Fig.\ref{overview}). The annotations were initially prelabeled by an algorithm developed by EYYES GmbH. Later, they were reviewed and refined manually. There is no annotation for the point cloud data.
\begin{figure}[ht]
  \centering
\includegraphics[width=1\linewidth]{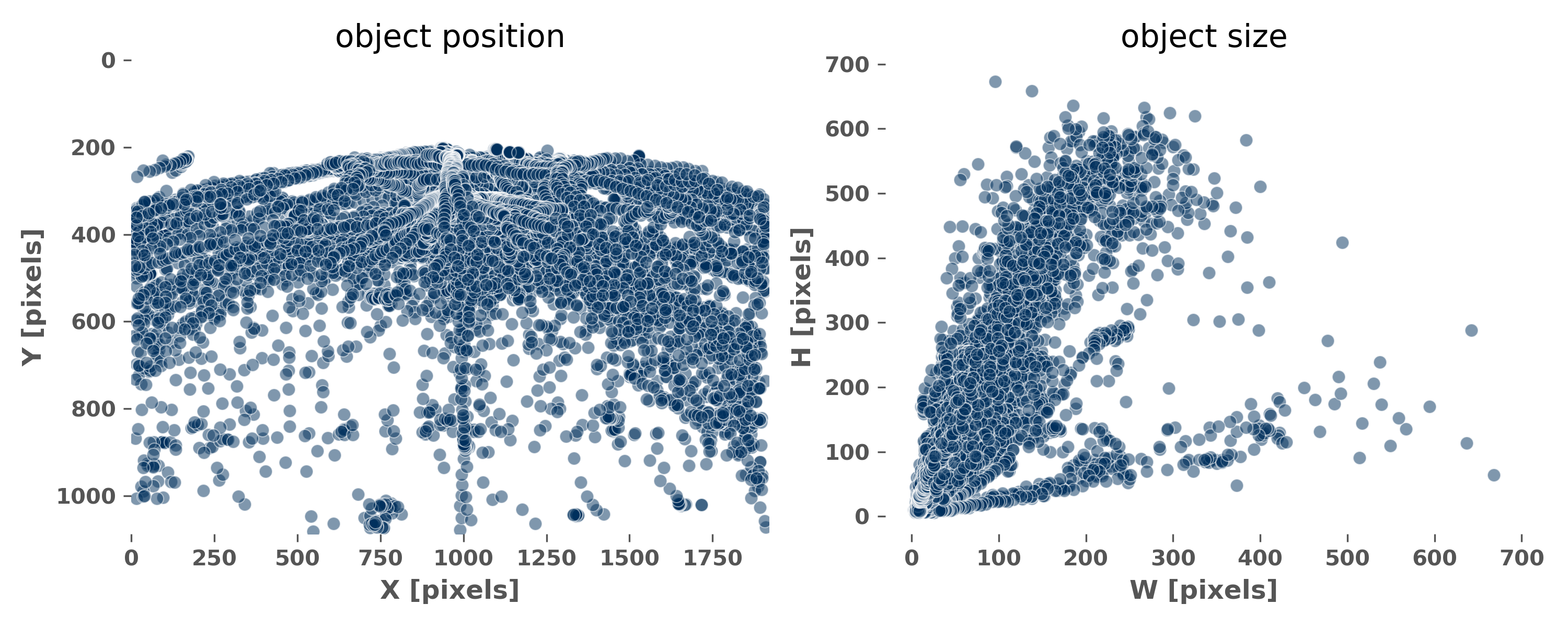}
\caption{Scatter plots of annotations. The left plot shows the distribution of object positions in the camera image, the right the distributions of height $H$ and width $W$.}
 \label{statistics}
\end{figure}
\section{\railgoerl~Statistics}
\label{acf}
\label{dss}

The annotated dataset comprises $61$ video sequences. These sequences include $\numprint{12205}$ frames with $\numprint{33556}$ bounding box annotations of persons. The video sequences were recorded at frame rate of $\qty{25}{\Hz}$ and every $\nth{15}$ frame starting from frame $0$ was extracted for annotation purposes. Annotations' scatter plots are depicted in Fig.\ref{statistics} showing a broad range of various person sizes as well as person positions. There were $6$ scenes with people lying down, $2$ fell, 2 between rails, $1$ next to the rails and $1$ across the rails.
\section{Limitations}
\label{Limitations}
In contrast to human vision, \railgoerl~is not a stereo data set as it contains multiple sensor types. Additionally, it cannot be used to train a sequence-based CV for a forward-moving train scenario, since due to safety reasons, actors were recorded from a reversing train. However, it can be used to evaluate the safety of models for hazard detection using XAI. Further, night-light recordings are not included. The fact that the recording area was a German rail test center limits its applicability to more general settings.  

\section{Conclusion and Future Work}
\label{futurework}
As trespassing on railways is life-threatening and disrupts train operations, person detection is critical. \railgoerl~is one of the $5$ open annotated datasets recorded on-board for mainline railways, which also can be used for urban railways. Further datasets similar to \railgoerl~will be required to develop CV systems enabling the development of GoA3+. They can include additional sensors and an increase of data in sense of quantity. \railgoerl~can serve as a reference as well as a basis for extensions, which will fill the gaps described in Sec.\ref{Limitations}. 

Developers of CV systems that work in related fields of research like automated security surveillance performed on-board and off-board might also take advantage of \railgoerl~and contribute to the development of further datasets. Similarly, developers of data generation systems \cite{damico2023trainsim} for railways  might also improve the performance of their system by using \railgoerl. All stakeholders in the rail sector and beyond are invited to participate in this effort and, if possible, publish new datasets to achieve a broad research and development community.%
\section*{ACKNOWLEDGMENT}
For their support in the project, the authors thank their DZSF colleagues K. Hofmann, K. Mühl, all amateur actors and manual data annotation correctors. This work was funded by the DZSF as part of the \href{https://www.dzsf.bund.de/SharedDocs/Standardartikel/DZSF/Projekte/Projekt_151_Sicherheitsargumentation_KI_ATO.html}{XRAISE project}, in-house research within the \href{https://www.bmdv-expertennetzwerk.bund.de/EN/Home/home_node.html}{BMDV Network of Experts} and also partly by the \href{https://secai.org/}{SECAI} (BMBF Project Nr. 57616814).
\bibliographystyle{IEEEtran}
\bibliography{railgoerl24}
\end{document}